\documentclass[11pt,a4paper]{article}
\usepackage[hyperref]{naaclhlt2018}
\usepackage{times}
\usepackage{latexsym}
\usepackage{amsmath}
\usepackage{graphicx}
\usepackage[utf8]{inputenc}
\usepackage[polish]{babel}

\DeclareMathOperator*{\argmax}{arg\,max}
\DeclareMathOperator*{\softmax}{softmax}
\usepackage{url}

\aclfinalcopy 

\title{TMLab SRPOL at SemEval-2019 Task 8: Fact Checking in Community Question Answering Forums}

\author{Piotr Niewiński, Aleksander Wawer, Maria Pszona, Maria Janicka\\
 Samsung R\&D Institute Poland \\
 pl. Europejski 1 \\
 00-844 Warsaw, Poland \\
 {\tt \{p.niewinski, a.wawer, m.pszona, m.janicka\}@samsung.com} \\
}

\date{}

\begin{document}
\maketitle
\begin{abstract}
The article describes our submission to SemEval 2019 Task 8 on Fact-Checking in Community Forums. The systems under discussion participated in Subtask A: decide whether a question asks for factual information, opinion/advice or is just socializing. Our primary submission was ranked as the second one among all participants in the official evaluation phase. The article presents our primary solution: Deeply Regularized Residual Neural Network (DRR NN) with Universal Sentence Encoder embeddings. This is followed by a description of two contrastive solutions based on ensemble methods.
\end{abstract}

\section{Introduction}
Community question answering forums are good platforms for knowledge sharing; hence, they are widely used sources of information. The growing popularity of such knowledge exchange leads to a growing need to automate the process of verifying the post quality. The first step, often overlooked, is to categorize each question and establish what kind of information the user seeks. 

Question classification has been mainly used to support question answering systems. Two main method types have been proposed in the literature: (1) rule-based approaches with linguistic features \cite{tomuro2004question, huang2008question, silva2011symbolic}, and (2) machine learning approaches \cite{zhang2003question, metzler2005analysis}. These methods are rather simple, due to the fact that question classification is often just a preprocessing step in a larger task. However, we can observe some recent advances in this area, such as ULMFiT \cite{howard2018universal}, which achieves state-of-the-art performance on the TREC dataset \cite{voorhees1999trec}.    

The present article describes our systems submitted to the SemEval 2019 competition Task 8 subtask A on question classification. The competition data set consisted of QatarLiving forum questions classified as FACTUAL, OPINION or SOCIALIZING. The training data contained only 1,118 questions. Moreover, according to our evaluation, human-level accuracy on this data set was about 0.75, which was relatively low. Therefore, we approached the task as a challenging classification problem.

The article is structured as follows. Section \ref{sec:preprocessing} presents our experiments with preprocessing methods. Section \ref{sec:submission} describes our official submission, where we propose an architecture utilizing several regularization methods to address the problem of the small data set. For comparative purposes, section \ref{sec:contrastive} presents two ensemble models as contrastive examples. Section \ref{sec:results} provides the results achieved by the models. Lastly, section \ref{sec:conclusions} concludes the discussion.

\section{Data Preprocessing}
\label{sec:preprocessing}

We tested a few simple text preprocessing setups. Unfortunately, none of them helped the models achieve improved results. Hence, they are here presented as negative results, and for reference.

First, all emojis were removed from the text, and all URLs were replaced with the string `url link'. Next, all dates and hours were replaced with `date' and `hour', respectively. Ordinal numbers -- i.e. $1^{st}$, $2^{nd}$, $5^{th}$ etc. -- were replaced with `nth', while the remaining numbers were substituted wtih `num'. All of these sequences were found using regular expressions. Furthermore, if most of the letters were uppercase, the whole text was lowercased.

Second, some of the forum-specific jargon was replaced with more generally used terms. This was achieved by an internally prepared dictionary that translated `qar' into `Qatar currency', `qling' into `browsing Qatar forum', `ql' into `Qatar forum', `villagio' to `Qatar shopping center', etc. Additionally, it helped us to correct common spelling errors, such as `doha' for `Doha' and `qatar' for `Qatar'. Finally, spelling correction was performed by a custom character-based CNN language model. This way, we hoped to obtain a better representation of texts when embedded into vectors.

However, the experiments showed that none of these methods brought significant improvement in classification accuracy. It seemed that noise removal, combined with text normalization, deprived the data of significant features and information which carried crucial meaning for preparing text embeddings. Therefore, we finally did not perform any preprocessing and worked on raw question subjects and body text.

\section{Primary Submission}
\label{sec:submission}

\subsection{Features}
The feature space for the models was created by combining three different sources of information:
\begin{enumerate}
    \item \emph{Universal Sentence Encoder} -- The concatenated post subject and body text were embedded with the Universal Sentence Encoder (USE) \cite{cer2018universal} to create a 512-dimensional vector representation.
    \item \emph{fastText embeddings} -- The concatenated post subject and body text were tokenized with the Spacy library and embedded on the word level with 300-dimensional fastText vectors. Then, the vectors were averaged on the sequence dimension.
    \item \emph{Category statistics} -- For each QL post category, the ratio of the FACTUAL, OPINION and SOCIALIZING labels was calculated. The three numbers were normalized, forming a 3-dimensional vector.
\end{enumerate}

The three subfeature vectors were concatenated to produce an 815-dimensional vector for each question.

\subsection{Model Architecture}
\begin{figure}[h]
\includegraphics[width=0.4\textwidth]{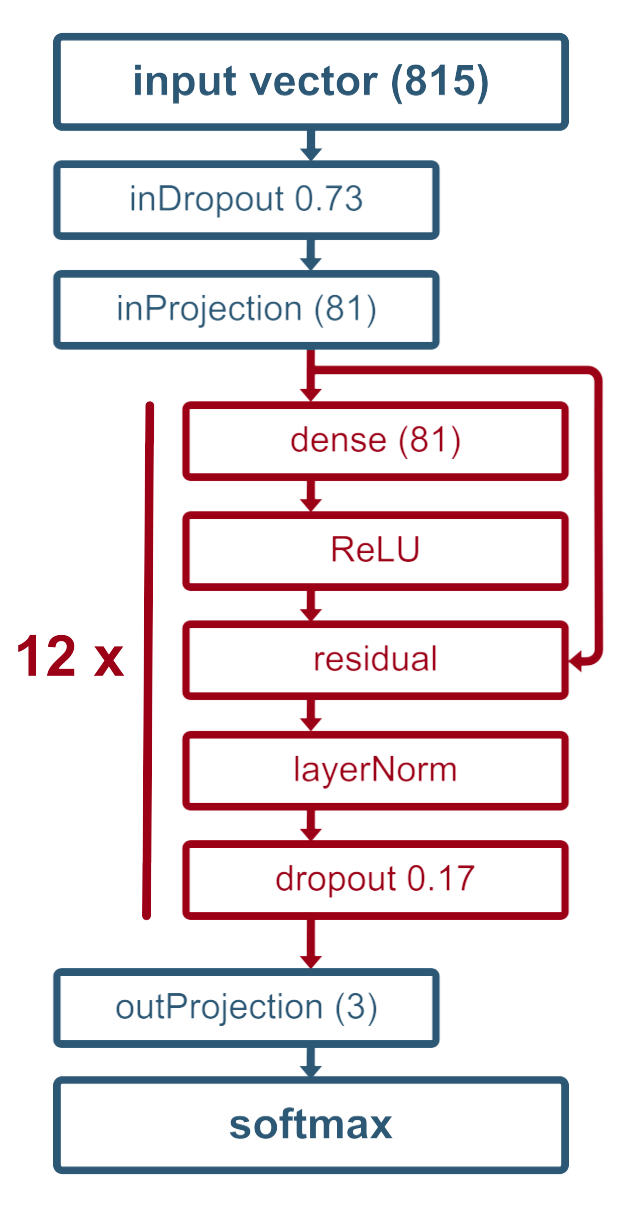}
\caption{The architecture of DRR NN (primary submission).}
\label{fig:model}
\end{figure}

We proposed the Deeply Regularized Residual Neutral Network architecture, shown in Figure \ref{fig:model}.

The model took as its input the 815-dimensional vector of floats (concatenated USE embeddings, fastText embeddings averaged, and category statistics). During the training, a large dropout of 0.73 was applied to the input vector. 

The core of the model was a deep subnetwork built of 12 stacked blocks. Each block contained an 81-dimensional dense layer followed by ReLU activation, residual connection, layer normalization and 0.17 dropout. Finally the output of the last block was projected with a dense layer into a 3-dim logits vector.

The model was trained with the Adam optimizer, at a 6e-3 learning rate, and with 500-epoch linear warmup. We used softmax cross entropy loss with 0.14 of L2 penalty regularization.

All model hyperparameters were optimized with a randomized search algorithm and 5-fold cross-validation over the training data set. The final model size was 148K learnable variables.

\subsection{Model Training}

The main idea behind the advanced training procedure was to split the training data into a bigger learning part and a smaller validation part. The loss was minimized on the learning part until the accuracy on the validation part began to increase.

Generally, model performance depends on many factors, such as training efficiency, model architecture, optimization algorithms, etc. At the same time, it is affected by sample distribution between learning and validation parts. 

In order to aggregate more knowledge from the training data, we used 5-fold cross-validation splits.
We prepared 4 such splits using different random seeds. This procedure gave us a total of 20 different pairs of learning/validation sets.

We set the maximum number of epochs to 700. The model was validated after each training epoch and saved until its classification accuracy improved. Usually, the accuracy was improving for about 300-600 epochs. For the final prediction, we used the argmax of the summarized softmax of 20 models:
\begin{equation*}
\argmax \sum_{k=1}^{20} \softmax(logits_k). 
\end{equation*}

\section{Contrastive Submissions}
\label{sec:contrastive}

For the contrastive submissions, our overall idea was to utilize multiple models that were as varied as possible, and combine their outputs.

In the first step, we used the following systems to obtain label probabilities for each  question:
\begin{itemize}
    \item \emph{ELMO} \cite{elmo2018} -- a deep, contextualized word representation to obtain sentence representation, followed by a neural network of two dense layers. We arrived at the following architecture and hyper-parameters during the optimization: a dense layer of 48 neurons (dropout 0.5), followed by a second dense layer of 10 neurons (dropout 0.5). When tested on the training data in cross-validation, this solution alone achieved a micro-accuracy of 0.72.
    
    \item \emph{BERT} \cite{bert2018} -- a deep, bidirectional transformer model with sequence classification layers on the top. The BERT language model was pre-trained, so only the sequence classifier was initialized and trained on the SemEval data. We used the PyTorch implementation of the case-insensitive `base' version\footnote{\url{https://github.com/huggingface/pytorch-pretrained-BERT}} with the optimal number of epochs (10) determined on the development set. When tested on the training data in cross-validation, this solution alone achieved a micro-accuracy of 0.717. 
    
    \item \emph{Bag-of-words} -- a machine learning solution based on character n-gram vectorization with TF-IDF weighting and a linear kernel SVM classifier. We used the implementation from the scikit-learn package \cite{scikit-learn}. When tested on the training data in cross-validation, this solution alone achieved a micro-accuracy of 0.699. 
    
\end{itemize}

In the second step, we prepared two different ensemble models combining the probability outputs from \emph{ELMO}, \emph{BERT}, \emph{Machine Learning} and \emph{DRR NN}.

The first contrastive submission (\textbf{Contrastive-1}) used the SVM classifier with linear kernel. 
The second contrastive submission (\textbf{Contrastive-2}) was designed as a bagging classifier of 10 estimators, each a voting ensemble of logistic regression, random forest and SVM with linear kernel. 

\section{Results}
\label{sec:results}

Table \ref{tab:results} contains the results of the evaluation on the official test set. The primary submission and both contrastive submissions were presented during the official phase of the contest. After the official competition, we tested additional solutions. Surprisingly, we achieved the best results with the SVM classifier (RBF kernel) on the USE embedding (\textbf{Post-evaluation}).

\begin{table}[!ht]
\begin{center}
\begin{tabular}{|l|ccc|}
\hline \bf Model & \bf Accuracy & \bf F1 & \bf AvgRec\\ \hline
DRR NN (Primary) & 0.83 & 0.72 & 0.76 \\
Contrastive-1 & 0.83 & 0.72 & 0.76\\
Contrastive-2 & 0.81 & 0.69 & 0.73\\
Post-evaluation & 0.87 & 0.77 & 0.78\\
\hline
\end{tabular}
\end{center}
\caption{Official results of our submissions on the test set. }\label{tab:results}
\end{table}

\section{Conclusions}
\label{sec:conclusions}

According to the experiments, and as reflected in the results on the test set, the best performing system was DRR NN based on the Universal Sentence Encoder. We attributed its good performance on the small data set to the deep regularization and the advanced training procedure. However, the SVM classifier performed even better, probably thanks to its overfitting  resistance \cite{xu2009robustness}.

Additionally, we tested several approaches, including the usual high performers, such as BERT or ELMO, and the ensemble systems. 
None of them was able to outperform our primary submission. We attribute such behaviour to data over-fitting and lack of ability to extract higher-level dependencies from the provided samples.

Some influence on the results could have been exerted by the significantly differing distributions of post categories among the train, dev and test sets. For example, while more than 30\% of all questions from the test set belonged to the `Visas and permits' category, only 8\% from the train set and 5\% from the dev set fall into the same category.

Linear SVM with the USE embeddings reached an accuracy of 0.84 on the dev set and 0.86 on the test set. Surprisingly, with a different set of parameters, we achieved 0.87 accuracy on the test set, and only 0.81 accuracy on the dev set.

\bibliography{semeval2018}
\bibliographystyle{acl_natbib}


\end{document}